\documentclass[journal]{IEEEtran}
\usepackage{amsmath,graphicx}
\usepackage{color,hyperref}
\usepackage[table,xcdraw]{xcolor}

\usepackage{times}
\usepackage{epsfig}
\usepackage{graphicx}
\usepackage{amsmath}
\usepackage{amssymb}

\usepackage{algorithm}
\usepackage{algorithmicx}
\usepackage{algpseudocode}
\usepackage{multirow}
\usepackage{booktabs} 
\usepackage{url}
\usepackage{bm}
\usepackage{verbatim}
\usepackage{breakurl}
\usepackage{bbding}
\usepackage{tabu}

\newcommand{\R}{\mathbb{R}}
\newcommand{\E}{\mathbb{E}}

\def\L{{\cal L}}

\newlength\savedwidth

\begin{document}
%
\title{ASK: Adaptively Selecting Key Local Features for RGB-D Scene Recognition}
\author{Zhitong~Xiong,
	Yuan~Yuan*,~\IEEEmembership{Senior Member,~IEEE,}
	and~Qi~Wang,~\IEEEmembership{Senior~Member,~IEEE}

\IEEEcompsocitemizethanks{
\emph{Corresponding author}: Yuan Yuan, y.yuan.ieee@gmail.com.
}
}

\markboth{IEEE TRANSACTIONS ON IMAGE PROCESSING}%
{Shell \MakeLowercase{\textit{et al.}}: Bare Demo of IEEEtran.cls for IEEE Journals}

\maketitle

\begin{abstract}
	Indoor scene images usually contain scattered objects and various scene layouts, which make RGB-D scene classification a challenging task. Existing methods still have limitations for classifying scene images with great spatial variability. Thus, how to extract local patch-level features effectively using only image label is still an open problem for RGB-D scene recognition. In this paper, we propose an efficient framework for RGB-D scene recognition, which adaptively selects important local features to capture the great spatial variability of scene images. Specifically, we design a differentiable local feature selection (DLFS) module, which can extract the appropriate number of key local scene-related features. Discriminative local theme-level and object-level representations can be selected with DLFS module from the spatially-correlated multi-modal RGB-D features. We take advantage of the correlation between RGB and depth modalities to provide more cues for selecting local features. To ensure that discriminative local features are selected, the variational mutual information maximization loss is proposed. Additionally, the DLFS module can be easily extended to select local features of different scales. By concatenating the local-orderless and global-structured multi-modal features, the proposed framework can achieve state-of-the-art performance on public RGB-D scene recognition datasets.
\end{abstract}

\section{Introduction}
\label{sec:intro}

Scene recognition is a fundamental task for computer vision. Recent progress made in deep convolutional neural networks (CNNs) has greatly boosted the performance of various computer vision tasks, such as image classification \cite{krizhevsky2012imagenet, yuan2018remote}, object detection \cite{ren2016faster,DBLP:journals/tip/YuanXW19}, semantic segmentation \cite{long2015fully, DBLP:journals/tip/WangGL19, yuan2019spatial} and video understanding \cite{DBLP:journals/tip/ZhaoLL19, gao2020feature, yuan2018structured} on large-scale benchmarks. Modern deep CNN architectures such as ResNet \cite{DBLP:conf/cvpr/HeZRS16} and DenseNet \cite{Li2017Classification} are well designed for high semantic-level image representation learning. However, directly applying these deep CNNs for scene recognition still suffers from a limitation: global image features are not flexible enough to represent the indoor scene image with cluttered objects and complex spatial layouts. Considering the difference between image classification and scene recognition, Zhou et al. \cite{DBLP:conf/nips/ZhouLXTO14} released a large scale scene classification dataset named \emph{Places}. They showed the effectiveness of pre-training CNN parameters on Places instead of the object-centric dataset \emph{ImageNet}.

\begin{figure}
	\centering
	\includegraphics[width=0.5\textwidth]{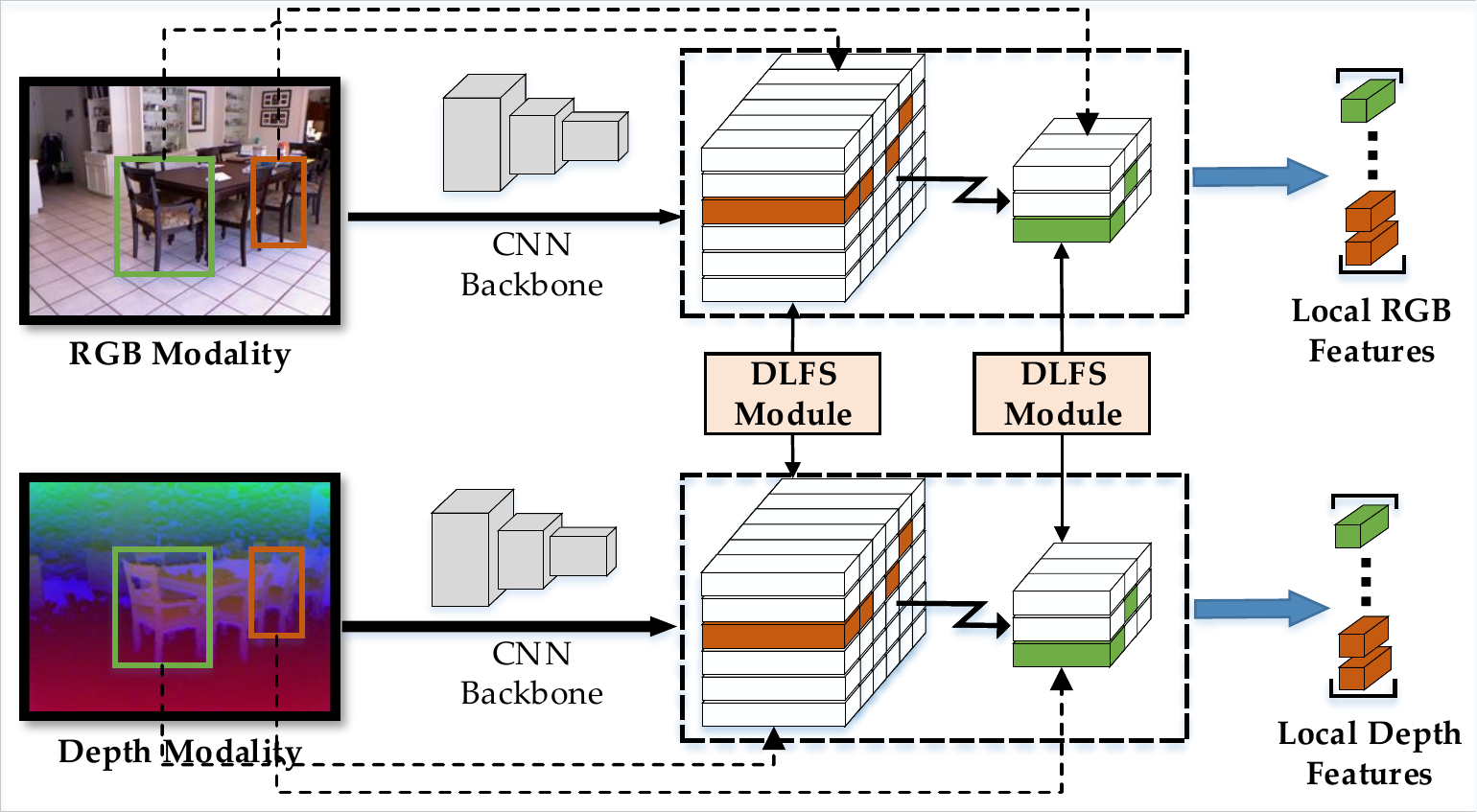}
	\caption{Motivation of the proposed method. We aim to select local object-level feature vectors from multi-modality deep intermediate feature maps in an unsupervised manner.}
	\label{motivation}
\end{figure}

\begin{figure*}
	\centering
	\includegraphics[width=1\textwidth]{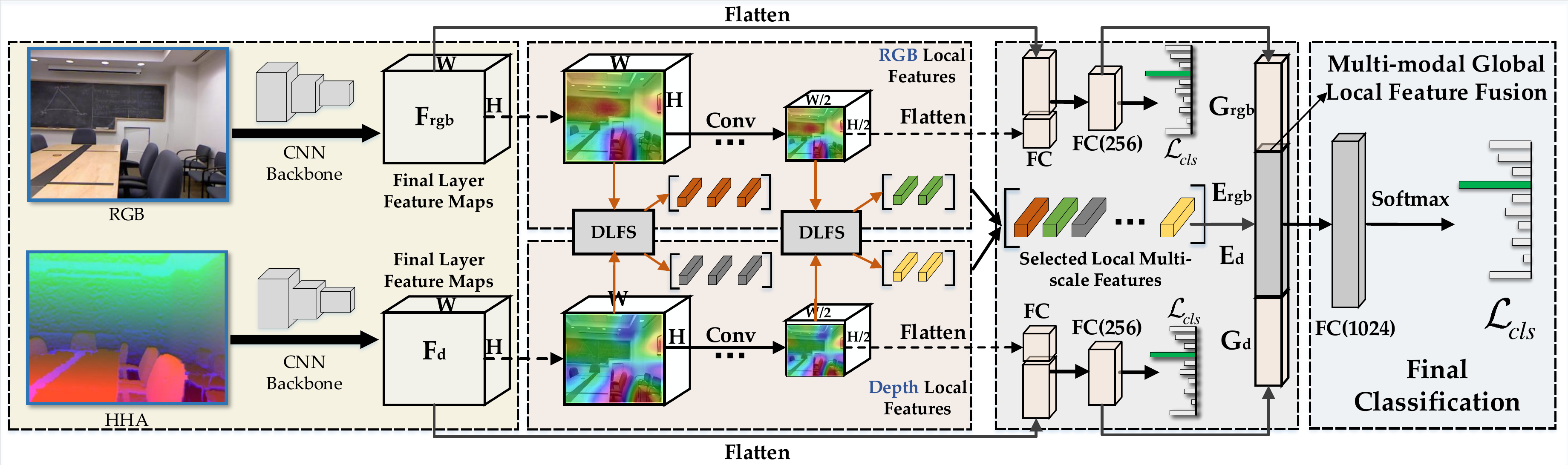}
	\caption{Main architecture of the proposed method. 1) \textbf{Global modal-specific feature learning.} The final layer feature maps of two modalities are input to two fully connected (FC) layers to learn global discriminative features with the supervision of auxiliary cross-entropy loss separately. 2) \textbf{Local modal-specific feature learning.} Meanwhile, the final layer feature maps are also used to construct a multi-scale feature pyramid, and then DLFS modules adaptively select multi-scale intermediate CNN features for scene recognition.}
	\label{arch}
\end{figure*}

The local object-level intermediate features are complementary to global CNN features \cite{DBLP:conf/eccv/KimF18}. Thus, selecting local features can be effective for taming the great geometric variability of scene image \cite{DBLP:journals/tgrs/YuanFLF19}. The local-feature based methods can be roughly divided into two categories. 1) Local patch-sampling based methods; 2) Object detection based methods. Several works \cite{DBLP:conf/eccv/GongWGL14,DBLP:journals/corr/YooPLK14,DBLP:conf/eccv/ZuoWSZYJ14} opted to extract features from different scales and locations densely and combined them via the fisher vector (FV) \cite{DBLP:journals/ijcv/SanchezPMV13}. However, two disadvantages exist in these methods, which limit the further performance improvement. The first one is that global layout features are neglected. The second one is that densely-sampled local features may induce scene-irrelevant noise, which makes the learned features less discriminative. Object detection can also be used to extract more accurate object-level local features. However, the performance of these methods greatly relies on the accuracy of object detection. Unfortunately, detecting the cluttered objects accurately in complex indoor scenes is also nontrivial.

Moreover, bottom-up local feature learning methods may also suffer from the following problems: Not all local object-level features contribute to the discriminative scene representations. For example, the `chair' features in dining room and the `chair' features in classroom are not discriminative for recognizing these two scenes. Merely using the local features may suffer from the ambiguity for recognizing different scenes. Some theme-level features are also important for scene classification, while object detection based methods may neglect them. For example, `floor' and `curtains' are background theme-level features, but they are also critical for recognizing scenes. 

Considering the aforementioned limitations, how to extract scene-related local features for RGB-D indoor image in a weakly supervised manner is still under explored. Since RGB-D image contains spatially-aligned multi-modality information, making use of the multi-modal feature learning process is helpful for local feature extraction. In this work, we find that the correlation between local objects of RGB and depth modalities is stronger than background regions. Therefore, exploiting the correlation distribution of multi-modality feature can help to extract semantic local features, which has not been investigated by previous works. Based on this finding, a differentiable local feature selection (DLFS) module is designed to adaptively extract deep intermediate local features. Additionally, modality correlation loss and mutual information maximization loss are proposed for training the DLFS module discriminately.

In this paper, we design a multi-modal feature learning framework for RGB-D scene recognition, which adaptively selects multi-scale key local features for scene recognition. As shown in Fig. \ref{motivation}, our motivation is to select multiple mid-level CNN feature vectors to represent local patches. To deal with the ambiguity problem caused by local features, the proposed framework also learns to extract global features which are important for describing the scene layout. The main contributions are as follows. 
\begin{enumerate}
	\item This work is the first to utilize the correlation between RGB and depth modalities to provide more cues for selecting local features. We design an effective differentiable local feature selection module based on the spatial-related correlation of multi-modal features.
	\item We introduce a mutual information maximization loss for training the DLFS module, which encourages the discrimination of selected local features.
	\item We design a compact global and local multi-modal feature learning framework to learn more discriminative representations for RGB-D scene recognition. 
\end{enumerate}

The rest of the paper is organized as follows. Related works about RGB-D scene recognition is shown in Section II. The detail of our method is presented in Section III. The experiments section IV demonstrated the performance of the proposed method. We compared the proposed method with its counterparts in detail in section V. Finally the conclusion of this paper is given in Section VI.

\section{Related Work}
\label{Related Work}
In this section, the related works will be reviewed briefly. For local feature learning methods, patch-based, object detection based and CNN intermediate feature based methods are summarized and reviewed. Moreover, multi-modality feature learning methods for RGB-D data are also surveyed.

Additionally, more detailed comparison between the proposed method and related works are provided in section \ref{cwem}.

\textbf{Patch-sampling based methods.} These approaches extract local features from the patch-based CNN intermediate representations. Gong et al. \cite{DBLP:conf/eccv/GongWGL14} designed a multi-scale CNN framework to sample the local patch features densely, and then encoded them via VLAD \cite{DBLP:conf/cvpr/JegouDSP10}. Some other methods \cite{DBLP:journals/corr/YooPLK14,dixit2015scene} represented the scene image with multi-scale local activations via the FV encoding. Depth image patches were exploited in the work of Song et al. \cite{song2017depth}. They first trained the model with densely sampled depth patches in a weakly-supervised manner, and then fine-tuned the model with the full image. Nevertheless, densely sampled patch features may contain noise, which limits the scene recognition performance. 

\textbf{Object-detection based methods.} To discard the irrelevant local features, several methods employed object detection for more accurate object-level features. Wang et al. \cite{wang2016modality} exploited the local region proposals to extract component representations, and encoded the local and global features together via fisher vector. Song et al. \cite{song2017rgb} employed Faster RCNN to detect objects on both RGB and depth images. More accurate object-level local features could be obtained with the object detection sub-module. They further modeled the object-to-object relation and achieved state-of-the-art scene recognition results \cite{DBLP:journals/tip/SongJWCC20}. Although improved performance could be obtained by \cite{wang2016modality,song2017rgb}, the two-stage pipeline methods suffered from the error accumulation problem. The higher computational complexity is also a limitation of these methods. Moreover, detecting small objects in clutter in indoor scenes is nontrivial itself.

Selecting intermediate CNN representations is useful for many applications. KeypointNet \cite{suwajanakorn2018discovery} presented an end-to-end geometric reasoning framework to learn latent category-specific 3D keypoints. KeypointNet can discover geometrically and semantically consistent keypoints adaptively without extra annotations. Zheng et al. \cite{zheng2017learning} proposed a multi-attention convolutional neural network, which consisted of convolution, channel grouping and part classification sub-networks. Wang et al. \cite{wang2018learning} showed that intermediate CNN representations could be enhanced by learning a bank of convolutional filters that captured class-specific discriminative patches without extra bounding box annotations. Xu et al. \cite{xu2018deep} proposed deep regionlets for object detection, which could select non-rectangular regions within the detection framework.

Multi-modality feature learning is critical for RGB-D scene recognition, and various methods have been proposed \cite{wang2018detecting}. Song et al. \cite{song2015sun} fused the two modal features by concatenating them to one fully connected layer. Wang et al. \cite{wang2015mmss} proposed to learn modal-consistent features between RGB and depth images. Li et al. \cite{LiZCHT18} learned the modal-consistent and modal-distinctive embeddings between two modalities simultaneously. Spatial correspondence of local objects in RGB and depth modalities was exploited by \cite{xiong2019rgb} for multi-modal learning. The work in \cite{du2019translate} employed cross-modal translation to explicitly regularize the training of scene recognition, which improved the generalization ability of the model.

\section{Our Method}
The whole network architecture of the proposed method is shown in Fig. \ref{arch}. Depth image is firstly transformed to HHA (Horizontal disparity, Height above ground, Angle of the pixel’s local surface normal with gravity direction) encoding \cite{gupta2014learning}. RGB and HHA images are input to two branches of CNNs for feature extraction. Then, the final layer feature maps of the two modalities are used for global and local multi-modal feature learning. Finally, the global and local features of two modalities are concatenated together to form the final scene representation.
\begin{figure}
	\centering
	\includegraphics[width=0.48\textwidth]{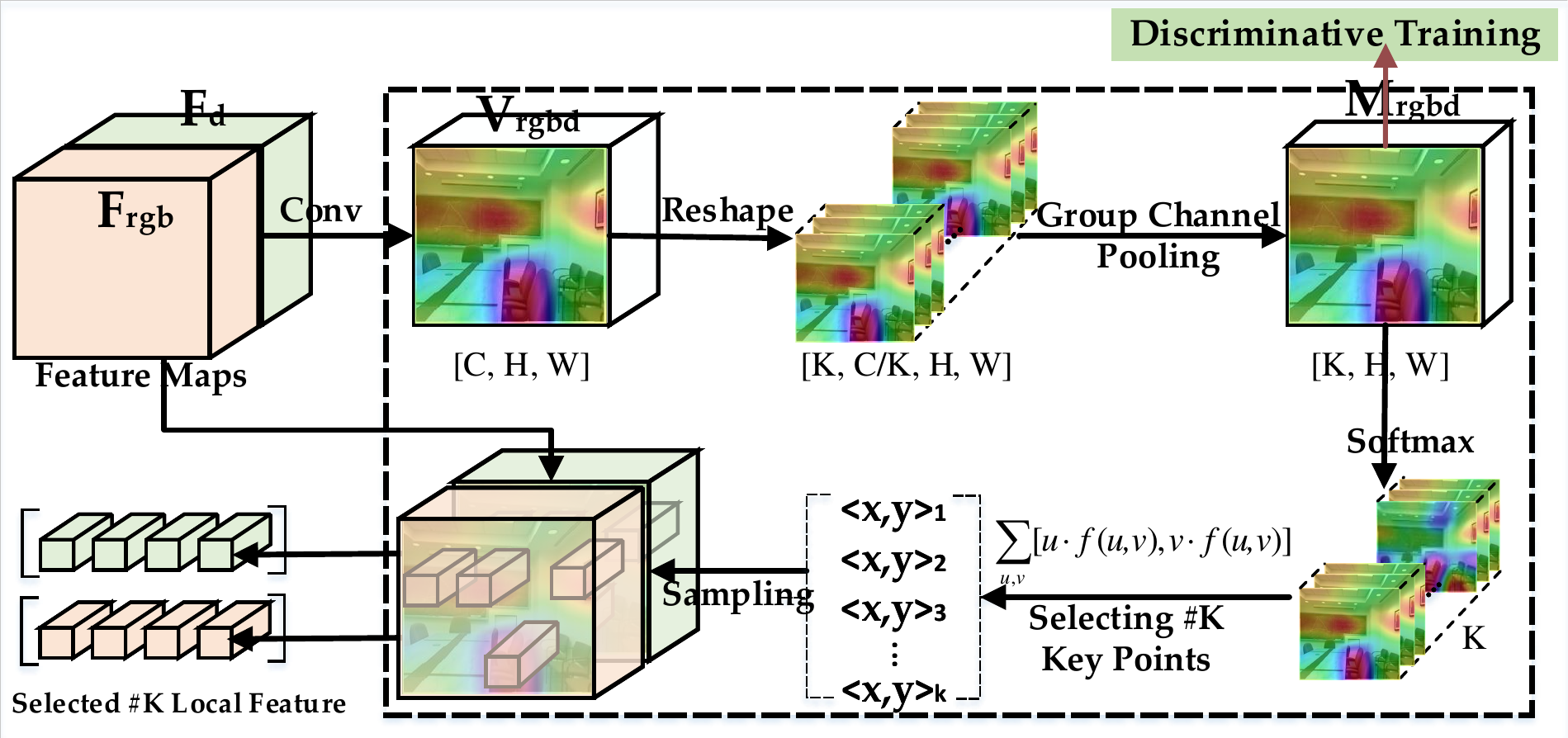}
	\caption{Illustration of the proposed DLFS module.}
	\label{ASKFS}
\end{figure}
\subsection{Differentiable Local Feature Selection}
CNNs can learn to extract high semantic-level features with stacked convolution layers. Layer visualization shows that the intermediate CNN features are with high abstract-level and can represent object parts. Moreover, the work of \cite{DBLP:conf/eccv/KimF18} find that the global scene feature and the object-level local feature are complementary for scene recognition. Based on this insight, our motivation is to select multiple $1\times 1$ mid-level CNN feature vectors to represent the local patches. Specifically, our model predicts $K$ keypoints with the final layer feature maps of RGB and depth modalities, as illustrated in Fig. \ref{ASKFS}.

Suppose that the final layer feature map of RGB and depth modalities are ${F_{rgb}}$ and ${F_{d}}$ respectively. We then concatenate these two feature maps as $F_{rgbd}$ to exploit multi-modal information for feature selection. The DLFS module takes ${F_{rgbd}}$ as input, and uses a $1\times 1$ convolution layer to learn the class-specific feature map for estimating $K$ keypoints. This can be formulated as 
\begin{equation}
{V_{rgbd}} = {conv_{1\times 1}(F_{rgbd})},
\end{equation}
where $V_{rgbd} \in \R^{(C,H,W)}$ is the discriminative feature map used for predicting keypoints. $C$ is the number of channels. $H$ and $W$ are the height and width of the feature maps respectively. ${conv_{1\times 1}}$ is a convolution layer with ${1\times 1}$ kernel size and output channel $C$. Although the channels of CNN feature maps are spatially-correlated, one separated channel is not enough to have strong part response \cite{zheng2017learning}. Thus, the channels of $V_{rgbd}$ are further grouped to be more spatially-correlated. Then we reshape feature ${V_{rgbd}}$ to ${V_{rgbd} \in \R^{(K,C/K,H,W)}}$ for the following group channel pooling operation. To make the part response stronger, we sum up the channel groups of ${V_{rgbd}}$ as
\begin{equation}
M_{rgbd}^j = \sum\limits_{i = 1}^{C/K} {{V_{rgb{d_i}}}},
\end{equation}
where ${M_{rgbd}\in \R^{(K,H,W)}}$ is the cross-channel grouped feature map for predicting $K$ selected keypoints. ${j \in \{1,2,...,K\}}$ is the index of the channel groups.

If there are no accurate keypoint annotations, directly training a mapping function from the input feature maps to keypoints is difficult. The reason is that the mapping model is hard to converge without keypoint annotations for supervised training. Inspired by \cite{suwajanakorn2018discovery}, we make our model output an attention map $h_j(u,v)$ to represent the probability of keypoint $j$ occurring at position $(u,v)$. In this work, we use the cross-channel grouped discriminative feature
maps $M_{rgbd}$ to output the attention map. Specifically, a 2D softmax layer is employed to produce the map $h$, which can be represented as
\begin{equation}
\begin{gathered}
h_j = softmax(M_{rgbd}^{j}) \in \R^{(H,W)},
\end{gathered}
\end{equation}
where ${h\in \R^{(K,H,W)}}$ is the probability distribution map, and $j$ is the index of predicted keypoints. Then the coordinates of the keypoints are computed by taking the expected values of the spatial distributions:
\begin{equation}
[{x_j},{y_j}] = \sum\limits_u^H {\sum\limits_v^W {[u \cdot {h_j}(u,v),v \cdot {h_j}(u,v)]} },
\end{equation}
where $[x_j, y_j]$ is the coordinate of the $j^{th}$ predicted keypoint.

To extract the local features in an end-to-end manner, the feature vector-sampling module should be differentiable. Without loss of generality, let ${E_{rgb} \in \R^{(K,C)}}$ be the selected local features for RGB modality. Similarly, we denote $E_{d}$ as the selected local features of the depth modality. Inspired by \cite{DBLPJaderbergSZK15}, the differentiable bilinear feature sampling can be formulated as
\begin{equation}
\begin{split}
E_{rgb_j}^c = \sum\limits_u^H \sum\limits_v^W F_{rgb}^c(u,v)  max(0,1 - |{x_j} - v|) \\ 
\cdot max(0,1 - |{y_j} - u|), 
\end{split}
\end{equation}
where ${j \in \{1,2,...,K\}}$ is the index of $K$ sampled local feature vectors, and ${c \in [1...C]}$ is the channel index. The coordinates ${(x_j,y_j)}$ and ${(u,v)}$ are normalized in the range of [-1,1].
\begin{figure}
	\centering
	\includegraphics[width=0.48\textwidth]{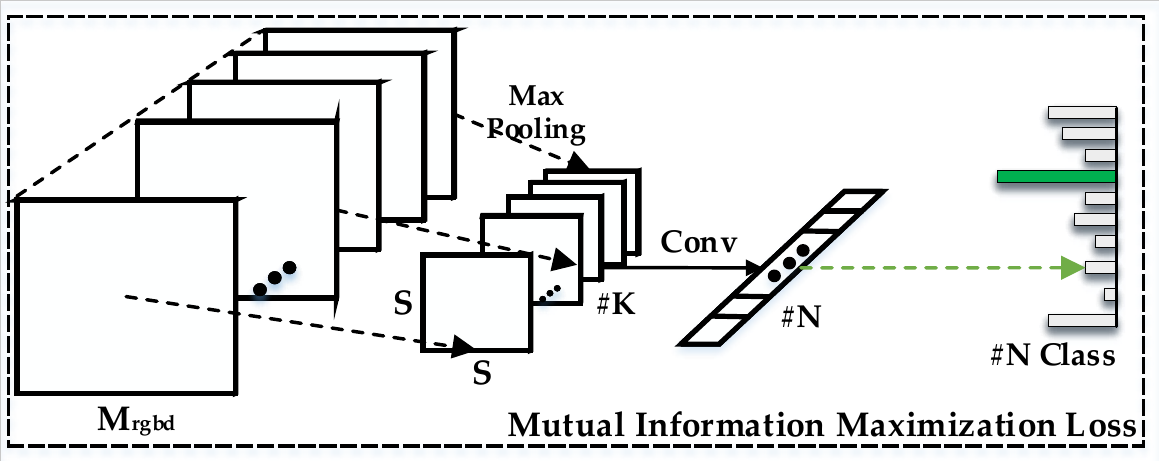}
	\caption{Illustration of the proposed variational information maximization loss.}
	\label{LOSS1}
\end{figure}
Since the DLFS module takes the keypoint index and the feature maps as input, we need to compute the partial derivatives of $F_{rgbd}$ as well as the predicted coordinates to allow the backpropagation through this module. The partial derivative w.r.t feature maps ${F_{rgb}}$ and $(x_j,y_j)$ are presented as follows.
The partial derivative of feature maps ${F_{rgb}}$ can be formulated as:
\begin{equation}
\begin{split}
\frac{{\partial E_{rgb}^c}}{{\partial F_{rgb}^c}} = \sum\limits_u^H \sum\limits_v^W max(0,1 - |{x_j} - v|) \\ 
\cdot max(0,1 - |{y_j} - u|).
\end{split}
\end{equation}
The partial derivatives of $y_j$ is
\begin{equation}
\label{e7}
\frac{{\partial E_{rgb}^c}}{{\partial {y_j}}} = \sum\limits_{u,v}^{H,W} F_{rgb}^c(u,v) {max(0,1 - |{x_j} - v|)g(v,{y_j})} ,
\end{equation}
where $g(v,y_j)$ is a piecewise function, and it can be formulated as
\begin{equation}
\label{e8}
g(v,{y_i}) = \left\{ {\begin{array}{*{20}{c}}
	{\begin{array}{*{20}{c}}
		0,\\
		{\begin{array}{*{20}{c}}
			1,\\
			{ - 1,}
			\end{array}}
		\end{array}}&{\begin{array}{*{20}{c}}
		{{\rm{if}}|v - {y_j}| \ge 1}\\
		{\begin{array}{*{20}{c}}
			{{\rm{if }}v \ge {y_j}}\\
			{{\rm{if }}v{\rm{  <  }}{{\rm{y}}_j}}
			\end{array}}
		\end{array}}
	\end{array}} \right..
\end{equation}

As for the depth modality ${F_{d}}$, the differentiable feature selection procedure is similar to the RGB modality. 

\begin{figure}
	\centering
	\includegraphics[width=0.46\textwidth]{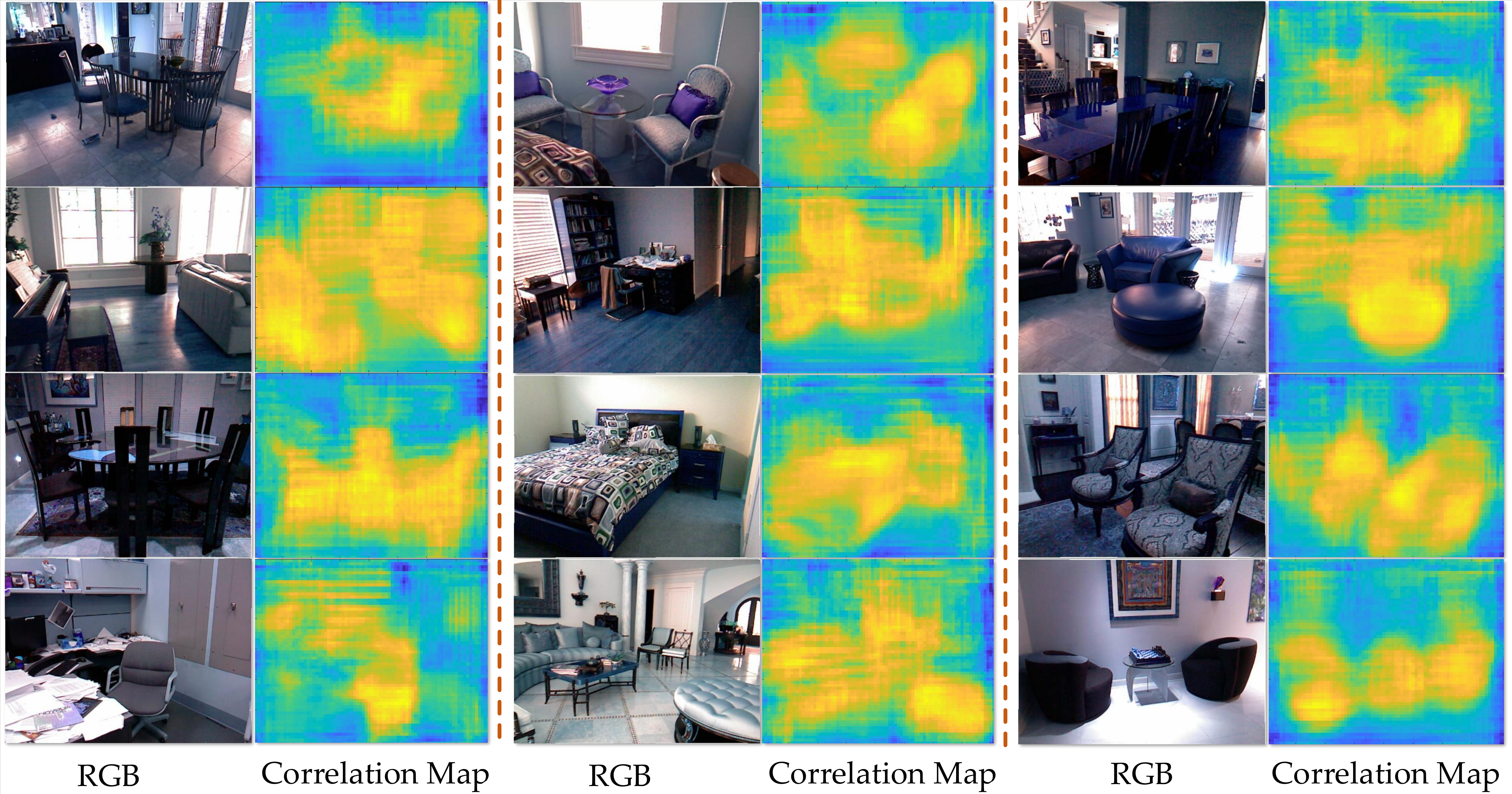}
	\caption{Pixel-wise correlation map between deep features of RGB and depth modalities. RGB images are on the left, and the corresponding correlation-maps are displayed on the right. More visualization examples are displayed in Fig. \ref{cosinesimilarity}.}
	\label{visfk}
\end{figure}
For multi-scale local feature selection, we can construct feature pyramid with stride 2 convolutional layers as shown in Fig. \ref{arch}. For each scale CNN feature maps, the local feature selection process is similar.

\subsection{Discriminative Training for DLFS}
Although the DLFS module can be trained in an end-to-end manner, the feature map ${M_{rgbd}}$ is not guaranteed to be discriminative enough for selecting different local patch features. To encourage the channels of $M_{rgbd}$ to be sensitive to different semantic parts, we propose a novel loss function to train the DLFS module. 
\begin{figure}
	\centering
	\includegraphics[width=0.48\textwidth]{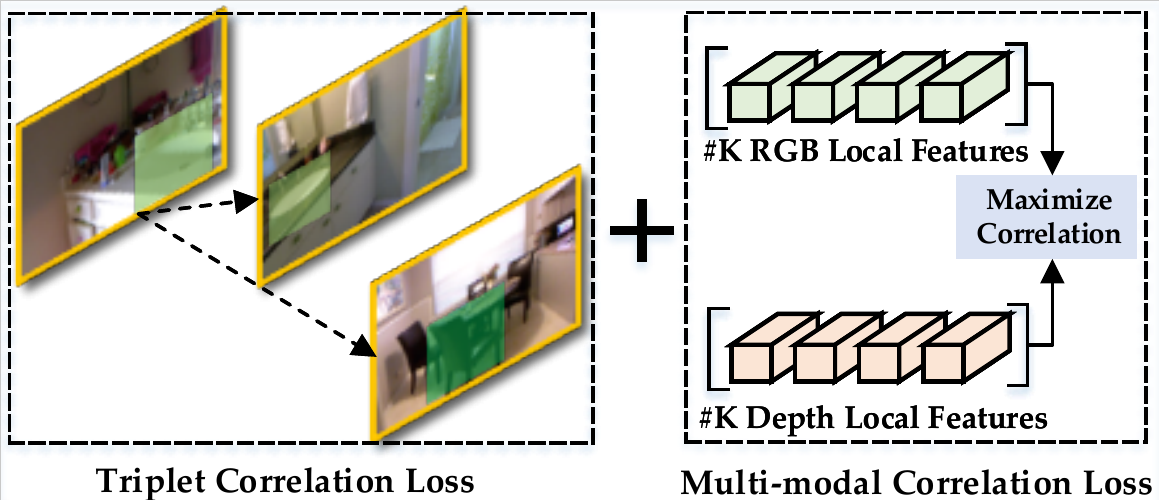}
	\caption{Illustration of the proposed loss function for training DLFS module. The first term is the triplet-correlation loss, and the second one is the multi-modal local feature correlation loss. }
	\label{LOSS}
\end{figure}

In order to make ${M_{rgbd}}$ more discriminative, we choose to retain high mutual information between ${M_{rgbd}}$ and the scene class label. Since we aim to select feature vectors which are important for classifying different scenes, the feature map ${M_{rgbd}}$ should be highly correlated to different scene classes. By maximizing the mutual information between ${M_{rgbd}}$ and the class label $L$, the channels of ${M_{rgbd}}$ is enforced to be class-correlated. As the exact computation of mutual information is intractable, variational lower bound is used for the approximation in this work. 

Suppose that there are $N$ classes, ${M_{rgbd}}$ is firstly pooled by a max-pooling layer to get $T_{rgbd} \in \R^{S\times S\times K}$. Then a $S\times S$ convolution layer is used to transform $T_{rgbd}$ into $N$ scalar features. Finally, we get $N$ features ${U \in \R^{N}}$ corresponding to $N$ scene categories. The mutual information ${I(L;U)}$ can be defined as 
\begin{equation}
\begin{aligned}
&I(L;U) = H(L) - H(L|U)\\
&= H(L) + \E_{L,U}[\log q(L|U)]+KL\\
&\ge H(L) + \E_{L,U}[\log q(L|U)],
\end{aligned}
\end{equation}
where ${H(\cdot)}$ is the entropy, and $\text{KL}$ denotes the Kullback-Leiber divergence. Maximizing the mutual information is equivalent to minimizing the following loss function $\L_{\text{VI}} = -\E_{L,U}[\log q(L|U)]$. $\L_{\text{VI}}$ can also be interpreted as the reconstruction error. In this work, we choose the variational distribution as a Gaussian distribution with heteroscedastic mean:
\begin{equation}
\begin{aligned}
&{\cal L}_{\text{VI}} =\sum_{n=1}^{N}\log\sigma_n + \frac{(L_n-{U_{\mu}}_n)^2}{2\sigma_n^2} + c,
\end{aligned}
\end{equation}
where $c$ is a constant, and $U_{\mu}$ is the transformed features by FC layer as shown in Fig. \ref{LOSS1}. $\sigma$ is a learnable parameter to be optimized.
By minimizing $\L_{\text{VI}}$, the feature ${M_{rgbd}}$ is encouraged to be discriminative for selecting key local features.

\subsection{Unevenly distributed Multi-modal Correlation}
We find that local objects contribute highly to the multi-modal correlation. To verify this, we compute the pixel-wise correlation map $P\in \R^{H,W}$ between the deep features $F_{rgb}\in \R^{C,H,W}$ of RGB modality and the depth modal deep features $F_{d}\in \R^{H,W}$. Then the pixel-wise correlation map can be defined as:
\begin{equation}
\begin{gathered}
P^{ij} = \rho({F_{rgb}}^{ij},{F_{d}}^{ij}), \\
i=1,...,H;j=1,...,W,
\end{gathered}
\end{equation}
where $\rho$ is the cosine similarity function. ${F_{rgb}}^{ij} \in \R^{C}$ is a feature vector at position $(i,j)$ of feature $F_{rgb}$.
As shown in Fig. \ref{visfk}, we have visualized the pixel-wise correlation map $P$. It can be clearly seen that the correlation between two modalities is unevenly distributed. More specifically, large correlation values cluster on local objects. This indicates that multi-modal features of local objects have higher correlation than other spatial positions. 

Based on this idea, local objects can be located by finding the largest correlation regions. Thus, we propose to maximize the correlation between RGB and depth local features to supervise the local feature selection module. The multi-modal correlation loss ${\cal L}_{C{m}}$ can be computed as:
\begin{equation}
{\cal L}_{C{m}} = \rho({E_{rgb}},{E_{d}}),
\end{equation}
where $\rho$ is the cosine similarity function. To further enhance the DLFS module with class-specific local features, triplet correlation loss is employed. As illustrated in Fig. \ref{LOSS}, we aim to select local features that have larger correlation between the same class and smaller correlation between different classes.

For each sample in the triplet input ${\{\textbf{a},\textbf{p},\textbf{n}\}}$, the corresponding selected local features are ${\{E_{rgb}a,E_{rgb}p,E_{rgb}n\}}$. $E_{rgb}p$ and $E_{rgb}a$ are positive and anchor features with the same class label, and $E_{rgb}n$ is the negative one with different scene class label. The triplet correlation loss is formulated as follows.
\begin{equation}
{\cal L}_{C{rgb}} = max\{ \rho({E_a},{E_p}) - \rho({E_a},{E_n}) + \alpha ,0\},
\end{equation}
where ${{\cal L}_{Crgb}}$ is the triplet correlation loss for the RGB modality, and the loss computation for depth modality ${{\cal L}_{Cd}}$ is similar. $\alpha$ is the margin value and it is set to 1.0 in this work. The whole correlation loss for two modalities can be formulated as ${{\cal L_C} = {\cal L}_{Crgb} + {\cal L}_{Cd}} + {\cal L}_{C{m}}$.

\subsection{Joint Global and Local Feature Representation}
Though local features are effective for representing the scene images, merely using local features may suffer from the ambiguity problem. Thus in this work, we choose to learn the global and local features simultaneously to obtain more robust representations for scene recognition.

The global features are learned by the FC layers connected to the final feature maps ${F_{rgb}}$ and ${F_{d}}$. As shown in Fig. \ref{arch}, two auxiliary cross-entropy loss functions are employed for learning global modal-specific feature separately. This loss function can be formulated as 
\begin{equation}
{{\cal L}_{aux}} = {{{\cal L}_{CE}}({G_{rgb}},{y})} + {{{\cal L}_{CE}}({G_{d}},{y})},
\end{equation}
where $L_{CE}$ represents the cross-entropy loss.

Finally, we concatenate the multi-modal global and local features together for the final scene classification. This can be denoted as 
\begin{equation}
\begin{gathered}
H_{mmgl}=concat(G_{rgb},G_{d},E_{rgb},E_d),\\
\end{gathered}
\end{equation}
where ${H_{mmgl}}$ is the multi-modal global and local feature vector. Then ${H_{mmgl}}$ is input to a fully connected layer, and the final classification result ${\hat y}$ is output through a softmax layer. We denote the final cross-entropy classification loss as ${{\cal L}_{cls}}$.

The overall loss function for training the proposed framework is a multi-task loss, which consists of  four terms: 1) the global modal-specific auxiliary loss ${\cal L}_{aux}$; 2) the mutual-information maximization loss ${\cal L}_{\text{VI}}$; 3) the triplet correlation and multi-modal correlation loss ${\cal L}_C$; 4) the final classification loss ${\cal L}_{cls}$. This can be represented as
\begin{equation}
\begin{aligned}
{\cal L} = {\cal L}_{cls} + \lambda_1 {\cal L}_{aux} + \lambda_2 {\cal L}_{\text{VI}} + \lambda_3 {\cal L}_C.
\end{aligned}
\end{equation}
where $\lambda_1$, $\lambda_2$, and $\lambda_3$ are balancing weights of loss components for the multi-task loss function.

\section{Experiments}
Two popular RGB-D scene recognition datasets are used to evaluate the proposed framework. One is the \textbf{SUN RGB-D} \cite{song2015sun} dataset, which contains 10,355 RGB and depth image pairs. They are captured from different depth sensors including Kinect v1, Kinect v2, Asus Xtion and RealSense. These images are divided into 19 scene categories. To compare with existing methods, we follow the same experimental settings with \cite{song2015sun}. For this dataset, 4,848 image pairs are used for training and 4,659 pairs for testing. Another dataset is the \textbf{NYUD v2} \cite{Silberman:ECCV12} dataset, which includes 1,449 RGB and depth image pairs divided into 10 categories. Following the experimental setting in \cite{gupta2013perceptual}, 795 image pairs are used for training and 654 pairs for testing.

\begin{table*}[h!]
	\centering
	\caption{Ablation Study on NYUD v2 Dataset}
	\label{tabel:nyu21}
	\scalebox{1.3}{
		\begin{tabular}{c|c|c}
			\hline \hline
			Feature Types &Methods & Mean-class Accuracy(\%) \\ \hline
			Single Modality & RGB / Depth / HHA & 61.2 / 54.1 / 58.2 \\ \hline
			\multirow{3}{*}{Multi-modality \& Global} 
			& RGB-D (HHA Encoding) & 64.2 \\ \cline{2-3}
			& RGB-D Global (${{\cal L}_{aux}}$) & 65.5 \\ \cline{2-3}
			& RGB-D Global (Spatial Attention \cite{Wang2018NonlocalNN}) & 66.1 \\ \hline
			\multirow{4}{*}{Multi-modality \& Local}
			& RGB-D Local  & {64.1} \\ \cline{2-3}
			& RGB-D Local (${\cal L_C}$) & {66.2} \\ \cline{2-3}
			& RGB-D Local (${\cal L_{\text{VI}}}$) & {66.7} \\ \cline{2-3}
			& RGB-D Local (Full DLFS) & {67.3} \\ \hline
			\multirow{4}{*}{Multi-modality \& Global \& Local}
			& RGB-D Global \& Local (${\cal L_C}$) & 67.1 \\ \cline{2-3}
			& RGB-D Global \& Local (${\cal L_{\text{VI}}}$) & 68.2 \\ \cline{2-3}
			& RGB-D Global \& Local (Full DLFS) & \textbf{68.9} \\ \cline{2-3}
			& RGB-D Global \& Local (Two-Scale Full DLFS) & \textbf{69.3} \\ \cline{2-3} \hline
		\end{tabular}
	}
\end{table*}

\subsection{Implementation Details}
HHA encodings is computed with the code released by \cite{gupta2014learning}. We use ResNet18 as the network backbone. Pre-trained parameters on \textbf{Places} are used for fine-tuning. Data augmentation is used in our work including random flip, cutout and random erasing \cite{zhong2017random}.

For optimization, Adam \cite{kingma2014adam} is employed with an initial learning rate of 1e-4, and the learning rate is reduced by a fraction of 0.9 every 80 epochs. The batch size is set to 64 with shuffle. For all the experiments, 300 epochs are used to train the model. As for the multi-task training, we set the parameters ${\lambda_1}$ to 1, and ${\lambda_2}$, ${\lambda_3}$ are set to 0.1 for all experiments. Multi-scale DLFS is evaluated with the ResNet18 backbone. The first scale feature maps are with the size of ${7\times7}$. The second scale is obtained with a $1\times1$ kernel convolution layer. By setting the stride to 2, the second scale feature maps are with the size of ${3\times3}$. For the first scale, $K$ is set to 16 and it is set to 4 for the second scale.

We randomly select 20\% training samples for each scene category to form a validation set for both datasets. With this setting, the dataset is split into training/validation/test sets. Training set is used for model training, and validation set for model selection. Then the test set is used for model evaluation and performance comparison.

\subsection{Ablation Study and Discussions}
To evaluate the proposed framework comprehensively, we do the following ablation studies with ResNet18 backbone to explore the effect of different sub-modules. Additionally, to show the superiority of multi-modality RGB-D data and local feature, we also conduct experiments to compare the performance of different feature types.

\subsubsection{\textbf{Single Modality}}
The results are shown in Table \ref{tabel:nyu21}. The first row displays the accuracy of using single modality: including RGB image, depth image and HHA image. ``RGB'' denotes RGB image. ``Depth'' denotes the original depth image, which is not transformed to HHA encoding. While ``HHA'' denotes that HHA image is used as depth modality input instead of the original depth image. Since RGB image contains richer texture and appearance information than depth modality, using single RGB modality data can achieve better accuracy. As for the depth modality, ``HHA'' can obtain better performance than ``Depth''. This indicates the effectiveness of HHA encoding.

\subsubsection{\textbf{Multi-modality \& Global}}
As shown in the table, ``RGB-D (HHA Encoding)'' is the method using global RGB and HHA features for scene classification, 
which can achieve obvious performance improvement compared with methods using single modality. This demonstrates that both RGB and depth modalities are useful for recognizing scenes. Additionally, we also study the effect of using auxiliary classification loss for multi-modality global features. ``RGB-D Global (${{\cal L}_{aux}}$)'' improves the baseline method by 1.3\%, which shows the effectiveness of the auxiliary loss for learning global modal-specific features.

Spatial attention mechanism can be used to focus on important local parts and extract more representative deep features.
In this work, we use the non-local neural networks \cite{Wang2018NonlocalNN} to focus on different spatial regions with assigned weights. Although spatial attention can be used to learn local-sensitive deep features, the final features of method ``RGB-D Global(Spatial Attention)'' are still global features processed by fully connected layer.
However, the proposed method selects key local features and abandon other features, which can be viewed as a kind of ``hard-attention". Using spatial-attention to focus on local features softly obtains a performance of 66.1\%, which is lower than the proposed local feature selection method.

\subsubsection{\textbf{Multi-modality \& Local}}
We have also evaluated the performance of the proposed method when only using local features extracted by DLFS module. As shown in Table \ref{tabel:nyu21}, by training DLFS module with the proposed loss functions ${\cal L_C}$ and ${\cal L_{\text{VI}}}$, ``RGB-D Local (Full DLFS)'' can achieve better results than the baseline methods. This demonstrates that multi-scale discriminative local features are useful for scene recognition. 

Since global RGB and depth features contain important scene layout information, which are complementary to local features. The proposed method exploits both local and global multi-modality features simultaneously, and can achieve better results than methods using global or local features alone.

\begin{figure}[!]
	\centering
	\includegraphics[width=0.48\textwidth]{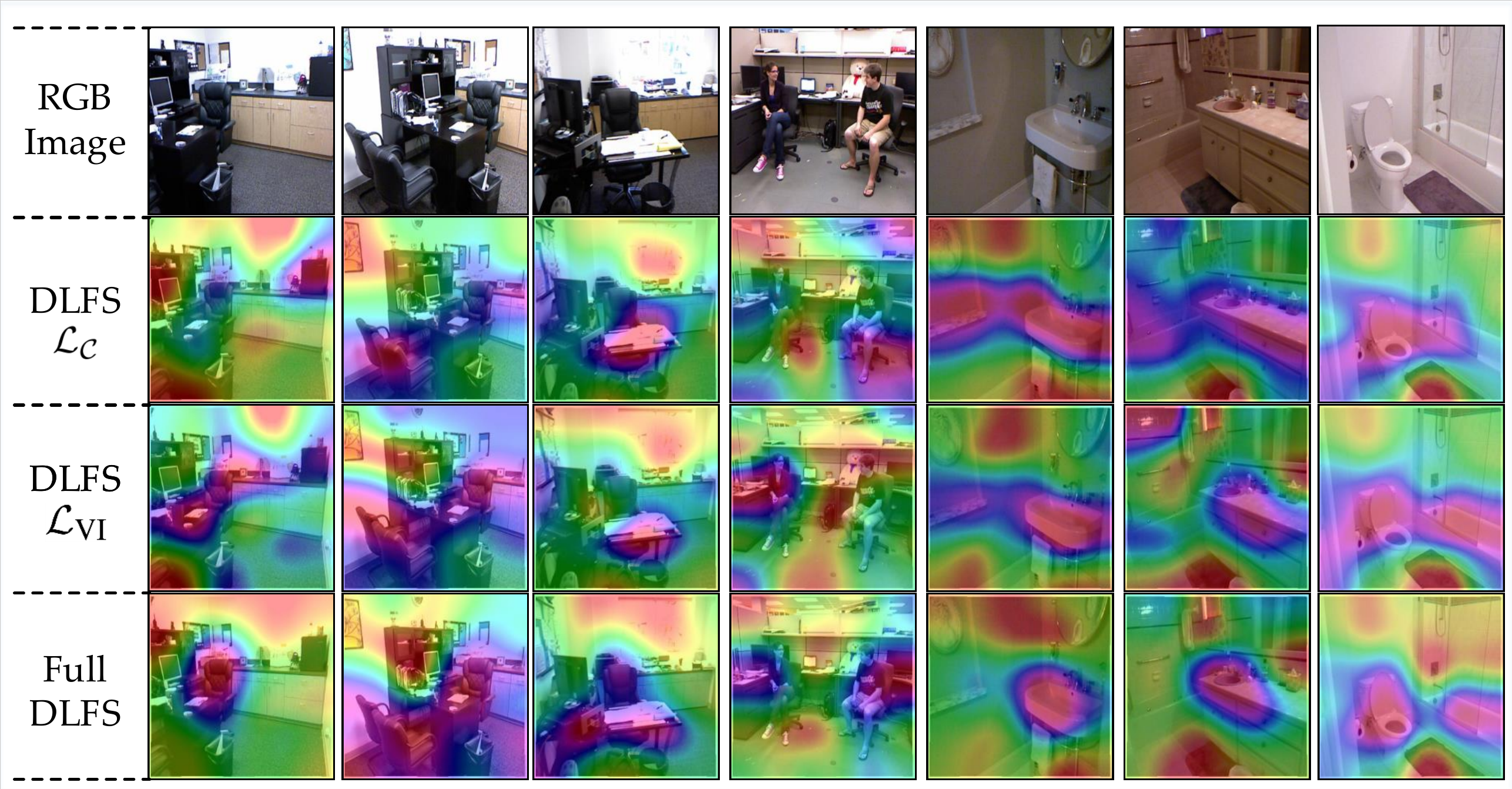}
	\caption{Class-specific activation map (CAM) \cite{DBLP:conf/cvpr/ZhouKLOT16} visualization of feature maps ${F_{rgb}}$. RGB images are shown in the first row. The second row shows the CAM of ``RGB-D Global \& Local(${\cal L_C}$)'', the third row shows the CAM of ``RGB-D Global \& Local(${\cal L_{\text{VI}}}$)'' and the fourth row shows the CAM of ``Full DLFS''.}
	\label{visf}
\end{figure}

\begin{table*}[h!]
	\centering
	\caption{Experimental Results on SUN RGB-D Dataset}
	\scalebox{1.26}{
		\label{tabel:sun1}
		\begin{tabular}{c|c|c|cc}
			\hline \hline
			Methods & Local Features & Multi-Modality Learning & Mean-class Accuracy(\%)  \\ \hline
			Song et al.\cite{song2015sun} & No & Feature-level Fusion & 39.0  \\ \hline
			Liao et al.\cite{liao2016understand} & No & Image-level Fusion & 41.3  \\ \hline
			Zhu et al.\cite{zhu2016discriminative} & No & \begin{tabular}[c]{@{}c@{}}Inter- \& Intra-\\ modality correlation\end{tabular} & 41.5  \\ \hline
			Wang et al.\cite{wang2016modality} & CNN Proposals & \begin{tabular}[c]{@{}c@{}}Local \& Global Features Fusion\end{tabular} & 48.1  \\ \hline
			Song et al.\cite{song2017rgb} & Object Detection & \begin{tabular}[c]{@{}c@{}}Local \& Global Features Fusion\end{tabular} & 54.0  \\ \hline
			Song et al. \cite{song2017combining} & No  & Global features  & 52.3  \\ \hline
			Song et al. \cite{song2017depth} & Local Patches & Local \& Global features Fusion  & 52.4 \\ \hline
			Li et al.\cite{LiZCHT18} & No & Modality Distinction \& Correlation & 54.6 \\ \hline
			Song et al.\cite{song2018learning} & Patches Sampling          & Feature-level Fusion                       & 53.8   \\ \hline
			Xiong et al.\cite{xiong2019rgb}    & Feature Selection           & Local \& Global features Fusion                        &55.9                           \\ \hline
			Song et al. \cite{song2019image}  & Object Detection & Local \& Global features Fusion & 55.5    \\ \hline
			ASK (K=16 \& K=4) & Local Feature Selection & \begin{tabular}[c]{@{}c@{}}Global \& Local Features\\ Modality Distinction  \end{tabular} & \textbf{57.3} \\ \hline
		\end{tabular}
	}
\end{table*}

\subsubsection{\textbf{Multi-modality \& Global \& Local}}
To study the effect of different loss functions of DLFS module, we do experiments to evaluate our framework with only  ${\cal L_C}$ and only ${\cal L_{\text{VI}}}$. As shown in the table, ``RGB-D Global \& Local (${\cal L_C}$)'' denotes the model with single-scale DLFS module and training loss ${\cal L_C}$. This indicates that using ${\cal L_C}$ to learn local features can improve the scene classification performance. The main reason is that local deep grid features are complement to global scene features. The loss ${\cal L_{\text{VI}}}$ is also effective for improving the performance, as it can encourage the selected local features to be correlated with the scene class. The results indicate that both loss functions are useful for improving the DLFS module.

Moreover, we have visualized the class-specific activation map of the learned features with ${\cal L_C}$ loss, ${\cal L_{\text{VI}}}$ loss and both losses, i.e. ``Full DLFS''. From Fig. \ref{visf} we can see that the features of ``Full DLFS'' are more spatially-correlated.

\subsection{SUN RGB-D Results}
The comparison results on SUN RGB-D dataset are displayed in Table \ref{tabel:sun1}. Among the compared methods, Song et al. \cite{song2015sun}, Liao et al. \cite{liao2016understand} and Zhu et al. \cite{zhu2016discriminative} did not use local features. Wang et al. \cite{wang2016modality} and Song et al. \cite{song2017rgb} employed object detection for scene recognition. 

Generally, the experimental results reveal that methods with local features can obtain better performance than those without local features. Different from existing methods, the proposed method selects multi-scale local features adaptively for scene recognition, and achieves even better performance.

We also summarize the multi-modality feature learning types in Table \ref{tabel:sun1}. Feature-level fusion are commonly used multi-modality feature learning methods, while considering the correlation and distinction between different modalities can achieve better results than simply combining multi-modality features. Different from existing methods, the proposed method exploits local features to learn modal-correlated representations. Additionally, the proposed method exploits the spatial-distribution of multi-modality feature correlation to enhance the local feature mining process. Meanwhile, the learned local features are also encouraged to be more modality-correlated by the proposed loss. By using this mechanism cleverly, the proposed framework with DLFS module achieves state-of-the-art scene recognition result.

\subsection{NYUD v2 Results}
Since NYUD v2 dataset is relatively small and the training data in NYUD v2 dataset is heavily imbalanced, we use the weights pretrained on SUN RGB-D dataset for model initialization on NYUD v2 dataset. The comparison results on NYUD v2 dataset are displayed in Table \ref{tabel:nyu1}. Similar to SUN RGB-D dataset, from the results we can see that methods using local features can achieve better performance. Although Li et al.\cite{LiZCHT18} did not use local features, they can still obtain competitive result by taking advantage of better multi-modality learning method. Compared with feature selection based method \cite{xiong2019rgb}, this work can achieve better performance by the differentiable feature selection module and the effective training loss functions.

\begin{figure}
	\centering
	\includegraphics[width=0.49\textwidth]{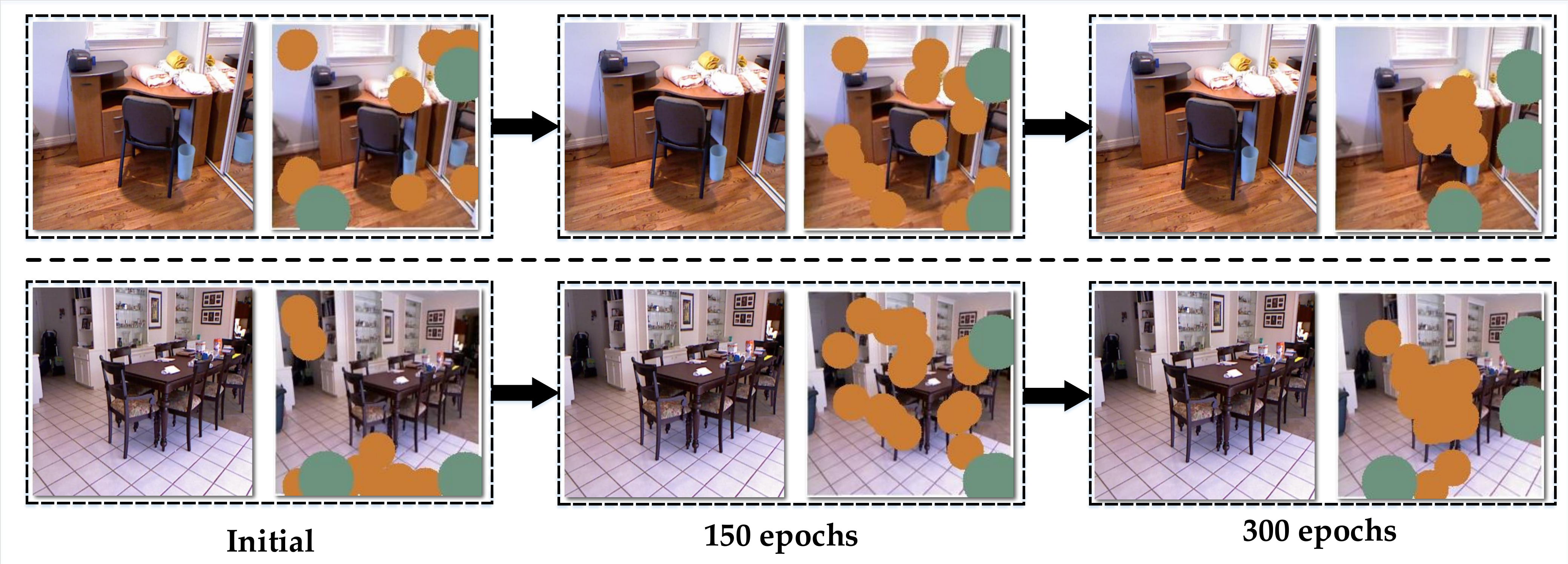}
	\caption{Illustration of the selected multi-scale keypoints during the training stage. Semantic-meaningful local features are selected when the modal converges. Different colors represent different scales. (Best viewed in color.)}
	\label{visp}
\end{figure}

The main advantages of the proposed method is two-fold: 1) local features extracted by DLFS module is more effective than patch-sampling and object detection methods; 2) Global modality-distinctive and local modality-correlated features are jointly exploited in this work. To sum up, the comparison results on this dataset indicate the effectiveness of the proposed framework. 
\begin{table*}[]
	\centering
	\caption{Experimental Results on NYUD v2 Dataset}
	\label{tabel:nyu1}
	\scalebox{1.33}{
		\begin{tabular}{c|c|c|c}
			\hline \hline
			& Methods                                              & Local Features                             &Mean-class Accuracy(\%)                     \\ \hline
			\multirow{8}{*}{State-of-the-art Methods} 
			& Gupta et al. \cite{gupta2015indoor} & No                                         & 45.4                           \\ \cline{2-4} 
			& Wang et al.\cite{wang2016modality} & CNN Proposals                              & 63.9                           \\ \cline{2-4} 
			& Li et al.\cite{LiZCHT18}           & No                                         & 65.4                           \\ \cline{2-4} 
			& Song et al.\cite{song2017depth}    & Local Patches                                         & 65.8                           \\ \cline{2-4} 
			& Du et al.\cite{du2019translate}  & No    & 66.5 \\ \cline{2-4}
			& Song et al.\cite{song2017rgb}      & Object Detection                           & 66.9 \\ \cline{2-4} 
			& Song et al.\cite{song2018learning} & Patches Sampling                                 & 67.5   \\ \cline{2-4} 
			& Xiong et al.\cite{xiong2019rgb}    & Feature Selection                                         & 67.8   \\ \cline{2-4} \hline
			{Proposed Method}                         & ASK (K=16 \& K=4)                                          & {Local Features Selection} & \textbf{69.3}                      \\ \hline
			
	\end{tabular}}
\end{table*}
Additionally, we have also done experiments to study the effect of selecting different number of local features. The results are displayed in Table \ref{sk}. \emph{`K=0'} denotes that no local features are used. Generally, the classification performance increases with more selected local features. However, when $K$ is set to larger than 16, the performance decreases. When we select more features (\emph{`K=36'}), the performance also decreases, which is only slightly better than \emph{`K=0'}. The reason is that when more local features are selected, more noise will also be introduced, and the global features may be suppressed. Object or theme-level features will be affected by the irrelevant noisy features.
\begin{table*}[]
	\centering
	\caption{Number of Selected Features on NYUD v2 Dataset}
	\scalebox{1.5}{
		\begin{tabular}{c|c|c|c|c|c|c|c}
			\hline \hline
			Number of Selected Features (K) & K=0 & K=1 & K=3 & K=9 & \textbf{K=16} & K=25 & K=36 \\ \hline
			Mean-class Accuracy (\%) & 65.4 & 66.2 & 67.1 & 67.9 & \textbf{69.1} & 67.7 & 66.1 \\ \hline
	\end{tabular}}
	\label{sk}
\end{table*}
The selected multi-scale keypoints are visualized in Fig. \ref{visp}. Different colors represent different scales. It can be clearly seen that semantic local features are selected when the modal converges.

\section{Comparisons with existing methods}
\label{cwem}
To show more details and give more comprehensive evaluations of the proposed work, we compare our work to other related existing methods and highlight the differences and similarities with them.
\begin{figure*}
	\centering
	\includegraphics[width=0.95\textwidth]{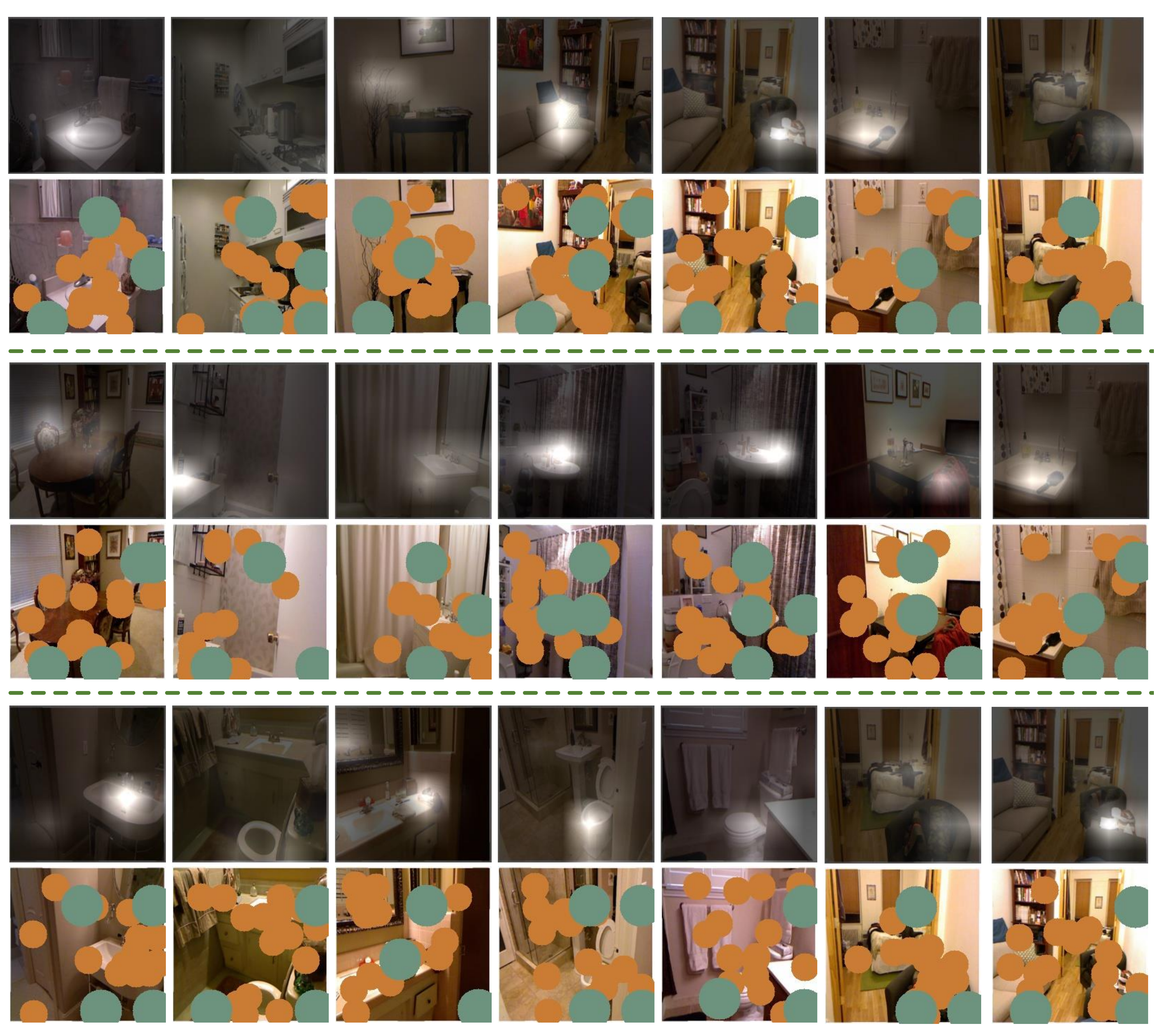}
	\caption{The attention map visualization of spatial attention based method \cite{wang2018non} and the selected keypoints visualization of the proposed method. The images of the first, third, fifth and seventh rows are the visualization for spatial attention based method. The second, fourth, sixth and eighth rows are the visualization of the selected keypoints of the proposed method. The upper and lower image pair is corresponding to the same image in NYUD v2 dataset \cite{Silberman:ECCV12}.}
	\label{vissao}
\end{figure*}
The highlight of the proposed framework is that it selects multi-scale intermediate CNN features instead of sampling patches densely or using the extra object detection procedure. Several other works have also proposed unsupervised local-part localization methods. Here we compare our method with them and point out the main differences with them.
\begin{figure*}
	\centering
	\includegraphics[width=0.92\textwidth]{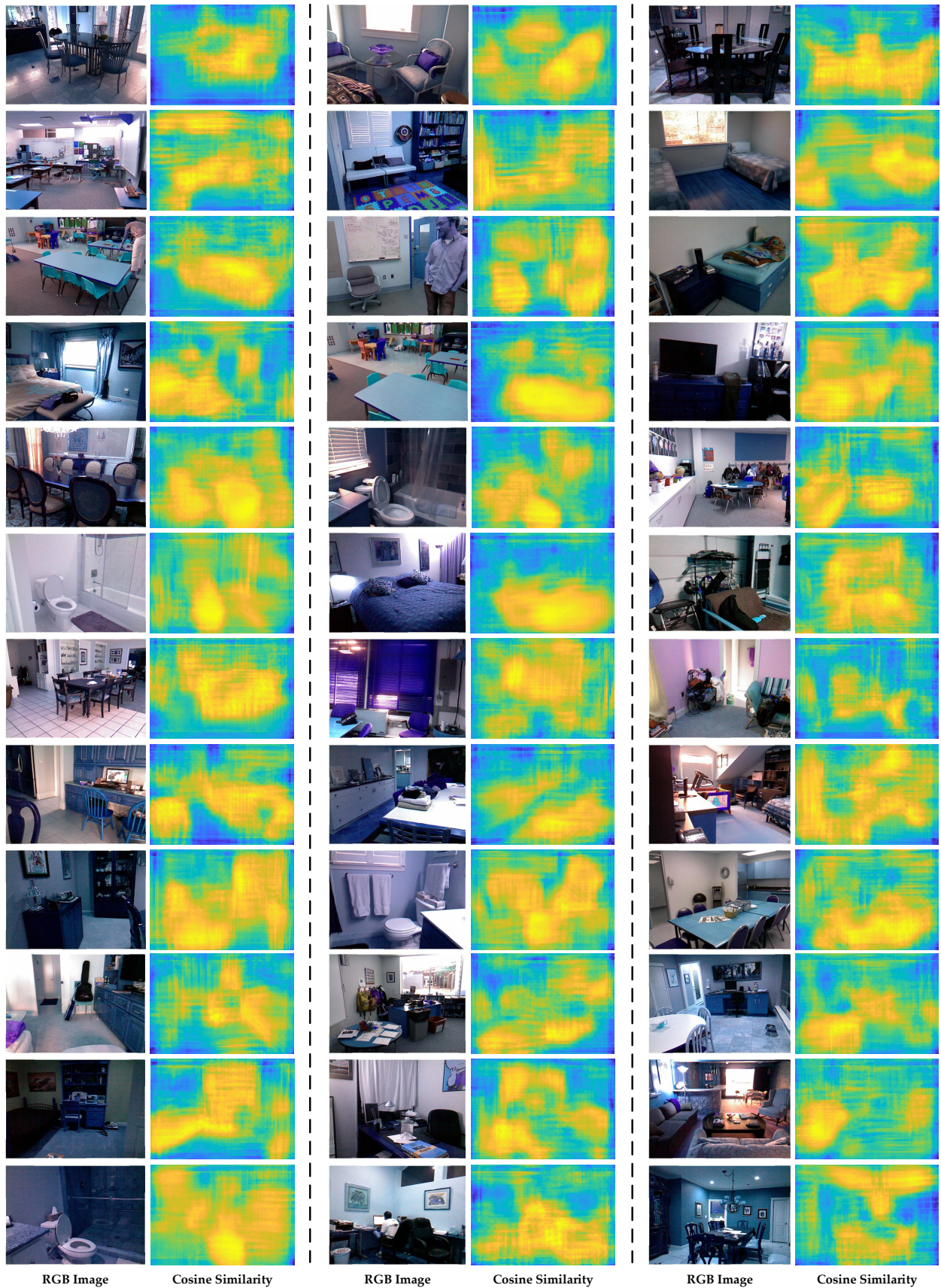}
	\caption{Spatial cosine-similarity map visualization. Images of the second, fourth and sixth columns visualize cosine-similarity maps.}
	\label{cosinesimilarity}
\end{figure*}

\textbf{Spatial-related Multi-modal Feature Learning} \cite{xiong2019rgb} found that similar spatial-attention maps are gained with attention mechanism for RGB and depth modalities, and enforcing similar spatial-attention map can boost the performance. This indicates that the same local objects are important features for both RGB and depth modalities. Inspired by this, we find that local objects contribute highly to the multi-modal correlation. Thus, we propose to maximize the correlation between RGB and depth local features, which provides more cues and supervision for local object-level feature selection. 

\textbf{KeypointNet \cite{suwajanakorn2018discovery}} KeypointNet is an end-to-end geometric reasoning framework for learning latent category-specific 3D keypoints. KeypointNet can discover geometrically and semantically consistent keypoints adaptively with no extra annotations. KeypointNet stacks 13 layers of dilated convolutions to output 2$N$ probability maps for predicting the 3$D$ coordinates of $N$ points. However, our work uses only one extra convolution layer with discriminative loss supervision for predicting probability maps, which is more time-efficient. Moreover, we use \emph{stride} 2 convolution layer instead of dilated convolution to select the multi-scale CNN features, which is also more time-efficient. Additionally, a novel variational mutual-information maximization loss term is proposed in this work for discriminative training.

\textbf{MA-CNN \cite{zheng2017learning}} MA-CNN consists of convolution, channel grouping and part classification sub-networks, which defines the parts as multiple attention areas. This work also employs the discriminative feature channels for part localization. However, the main difference is two-fold. 1) the local-part features are pooled from the attention areas in their work; 2) Multi-scale features are neglected. Different from their work, the proposed framework can learn to select multi-scale feature vectors in a totally differentiable manner.

\textbf{Deep Regionlets \cite{xu2018deep}} The architecture of deep regionlets includes a region selection network, which can learn more fine-grained features by selecting sub-regions adaptively. The local part feature selection is based on a spatial transformation and a gating network. STN \cite{DBLPJaderbergSZK15} is employed to select local regions as local-part features. However, the proposed method aims to select multi-scale mid-level CNN feature vectors (keypoints in CNN feature maps) as the local discriminative representations. 

\textbf{Discriminative Filter Bank \cite{wang2018learning}}  This work also exploits mid-level CNN representations by learning a bank of convolutional filters that capture class-specific discriminative patches without extra part or bounding box annotations. Specifically, this work uses Global Max Pooling (GMP) to select only one local feature vector from the intermediate CNN feature maps, which is different from our method. Moreover, multi-scale local features are also not considered in this work.

\textbf{Spatial Attention \cite{wang2018non}} To select important local features from mid-level CNN feature maps, spatial attention is the most commonly employed method. Although non-local networks \cite{wang2018non} can be used to focus on important local regions of feature maps, the softly attended feature maps still contain irrelevant features. However, our hard selection based method can select the important local feature vectors and discard the irrelevant ones. Moreover, the proposed method can select multiple multi-scale local features, which are more discriminative and representative than the spatial attention method. The visualization of the attention maps and the selected local feature vectors of our method are displayed in Fig. \ref{vissao}. The upper rows show the results of spatial attention based method, and the lower rows show the results of the proposed method. As we can see from Fig. \ref{vissao}, the spatial attention maps mainly focus on one local object regions, while our method can select multi-scale object-level (`chair' or `wash basin') and theme-level (`curtain' or `floor') feature vectors.

As displayed in Fig. \ref{cosinesimilarity}, it can be clearly seen that the correlation between RGB and depth modalities is highly spatially related. The correlation of local objects are higher than other positions.  Inspired by this intriguing finding, we can maximize the correlation between selected local multi-modal features to locate the positions of local objects.

\section{Conclusion}
In this work, we present a multi-modal global and local feature learning framework for RGB-D scene classification. The differentiable local feature selection (DLFS) module is proposed to select important local object and theme-level features adaptively for RGB-D scene images. A novel loss function is proposed to supervise the training of DLFS module. Discriminative local object and theme-level representations can be selected with DLFS module from the spatially-correlated multi-modal RGB-D features. We take advantage of the correlation between RGB and depth modalities to provide more cues for selecting local features. Additionally, we further enhance the DLFS module with the multi-scale feature pyramid to select object-level features of different scales.  Evaluations on SUN RGB-D and NYU Depth version 2 (NYUD v2) datasets have shown the effectiveness of the proposed framework.

\section{Acknowledgment}
This work was supported by the National Key Research and Development Project under Grant 2020YFB 2103902, National Natural Science Foundation of China under Grant 61632018, 61825603, U1801262, and 61871470.


\bibliographystyle{IEEEbib}
\bibliography{egbib}

\begin{IEEEbiography}[{\includegraphics[width=1in,height=1.25in,clip,keepaspectratio]{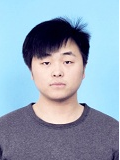}}]{Zhitong Xiong} received the M.E. degree in Northwestern Polytechnical University and is currently working toward the	Ph.D. degree with the School of Computer Science and Center for OPTical IMagery Analysis and Learning (OPTIMAL), Northwestern Polytechnical University, Xi'an, China. His research interests include computer vision and machine learning.
\end{IEEEbiography}

\begin{IEEEbiography}[{\includegraphics[width=1in,height=1.25in,clip,keepaspectratio]{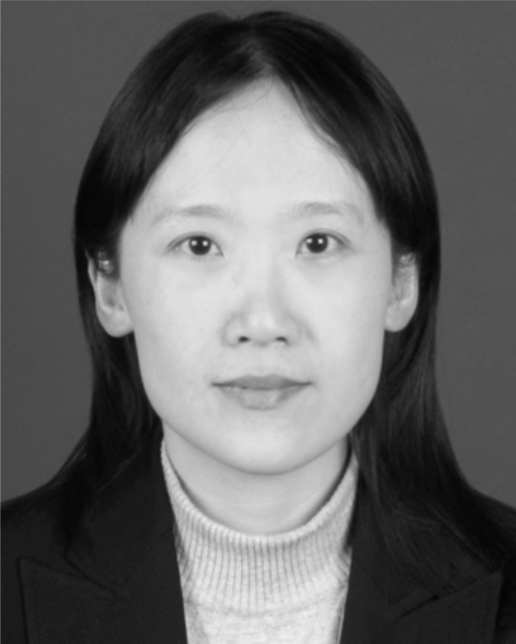}}]{Yuan Yuan} (M'05-SM'09) is currently a Full Professor with the School of Computer Science and the Center for Optical Imagery Analysis and	Learning, Northwestern Polytechnical University, Xi'an, China. She has authored or co-authored over 150 papers, including about 100 in reputable journals, such as the IEEE TRANSACTIONS AND PATTERN RECOGNITION, as well as the conference papers in CVPR, BMVC, ICIP, and ICASSP.	Her current research interests include visual information processing and image/video content analysis.
\end{IEEEbiography}

\begin{IEEEbiography}[{\includegraphics[width=1in,height=1.25in,clip,keepaspectratio]{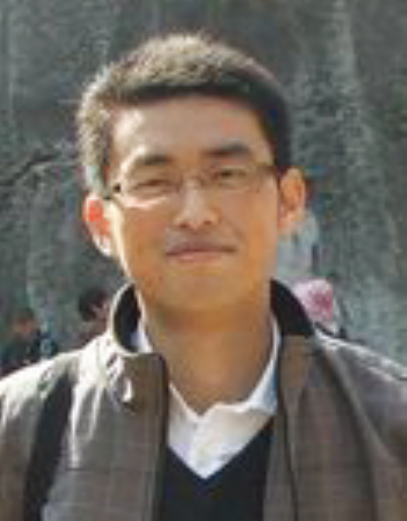}}]{Qi Wang} (M'15-SM'15) received the B.E.	degree in automation and the Ph.D. degree in pattern recognition and intelligent systems from the University of Science and Technology of China,	Hefei, China, in 2005 and 2010, respectively. He is	currently a Professor with the School of Computer Science and the Center for Optical Imagery Analysis and Learning, Northwestern Polytechnical University, Xi'an, China. His research interests include computer vision and pattern recognition.
\end{IEEEbiography}

\end{document}